%% file: emnlp2020.tex
\documentclass[11pt,a4paper]{article}
\usepackage[hyperref]{emnlp2020}
\usepackage{times}
\usepackage{latexsym}

\usepackage{microtype}

\usepackage{multirow}
\usepackage{amsmath}
\usepackage{{amsfonts}}
\usepackage{graphicx}
\usepackage{placeins}
\usepackage{wrapfig}
\usepackage{bm}
\usepackage{booktabs}
\usepackage{makecell}
\usepackage{xcolor}
\usepackage{url}

\aclfinalcopy % Uncomment this line for the final submission
\def\aclpaperid{3573} %  Enter the acl Paper ID here

\title{To Schedule or not to Schedule: Extracting Task Specific Temporal Entities and Associated Negation Constraints}

\author{Barun Patra, Chala Fufa, Pamela Bhattacharya, Charles Lee \\
  Microsoft \\
  {\tt \{bapatra, chfufa, pamelabh, charlle\}@microsoft.com}}

\date{}

\begin{document}
\definecolor{customOrange}{HTML}{ED7D31}
\definecolor{customGreen}{HTML}{548235}
\definecolor{customLightBlue}{HTML}{00B0F0}
\definecolor{customDarkBlue}{HTML}{002060}
\definecolor{customPurple}{HTML}{7030A0}

\maketitle
\input{texfiles/abstract}
\input{texfiles/introduction}
\input{texfiles/approach}
\input{texfiles/experiments}
\input{texfiles/results}
\input{texfiles/relatedWork}
\input{texfiles/conclusion}
\input{texfiles/acknowledgements}

\FloatBarrier

\bibliography{emnlp2020}
\bibliographystyle{acl_natbib}
\appendix
\clearpage
\input{texfiles/appendix}

\end{document}

%% file: texfiles/abstract.tex
\begin{abstract}
State of the art research for date-time\footnote{We use date-time entities, date entities, time entities and temporal entities interchangeably to denote entities associated with dates and/ or times.} entity extraction from text is task agnostic. Consequently, while the methods proposed in literature perform well for generic date-time extraction from texts, they don’t fare as well on task specific date-time entity extraction where only a subset of the date-time entities present in the text are pertinent to solving the task. Furthermore, some tasks require identifying negation constraints associated with the date-time entities to correctly reason over time. We showcase a novel model for extracting task-specific date-time entities along with their negation constraints. We show the efficacy of our method on the task of date-time understanding in the context of scheduling meetings for an email-based digital AI scheduling assistant. Our method achieves an absolute gain of 19\% f-score points compared to baseline methods in detecting the date-time entities relevant to scheduling meetings and a 4\% improvement over baseline methods for detecting negation constraints over date-time entities. 
\end{abstract}

%% file: texfiles/introduction.tex
\section{Introduction} 

Temporal entity extraction and normalization is an important aspect of Natural Language Processing \cite{alonso2011temporal, campos2014survey}. There has been a substantial body of work on the task and there exist numerous well performing publicly available models for identifying and normalizing temporal entities \cite{strotgen2010heideltime, chang2012sutime, zhong2018time}.  

There exist however a growing number of NLP applications which require extraction of only a relevant subset of time entities that are useful for solving specific problems within a larger body of text. Examples of such tasks include understanding search queries (``Find me all emails sent by April between May $11^{th}$ and May $21^{st}$''), Goal Oriented Dialogue Systems (``Deliver George Orwell's 1984 by next week.'', ``Send the ``FY 2020 Budget'' to Watson Monday morning.'') etc. Using the temporal entity extraction models for these tasks is insufficient, since they fail to disambiguate between general date-time entities and entities necessary to solve the task. 

\begin{figure*}[!thb]
    \centering
    \includegraphics[width=0.85\textwidth]{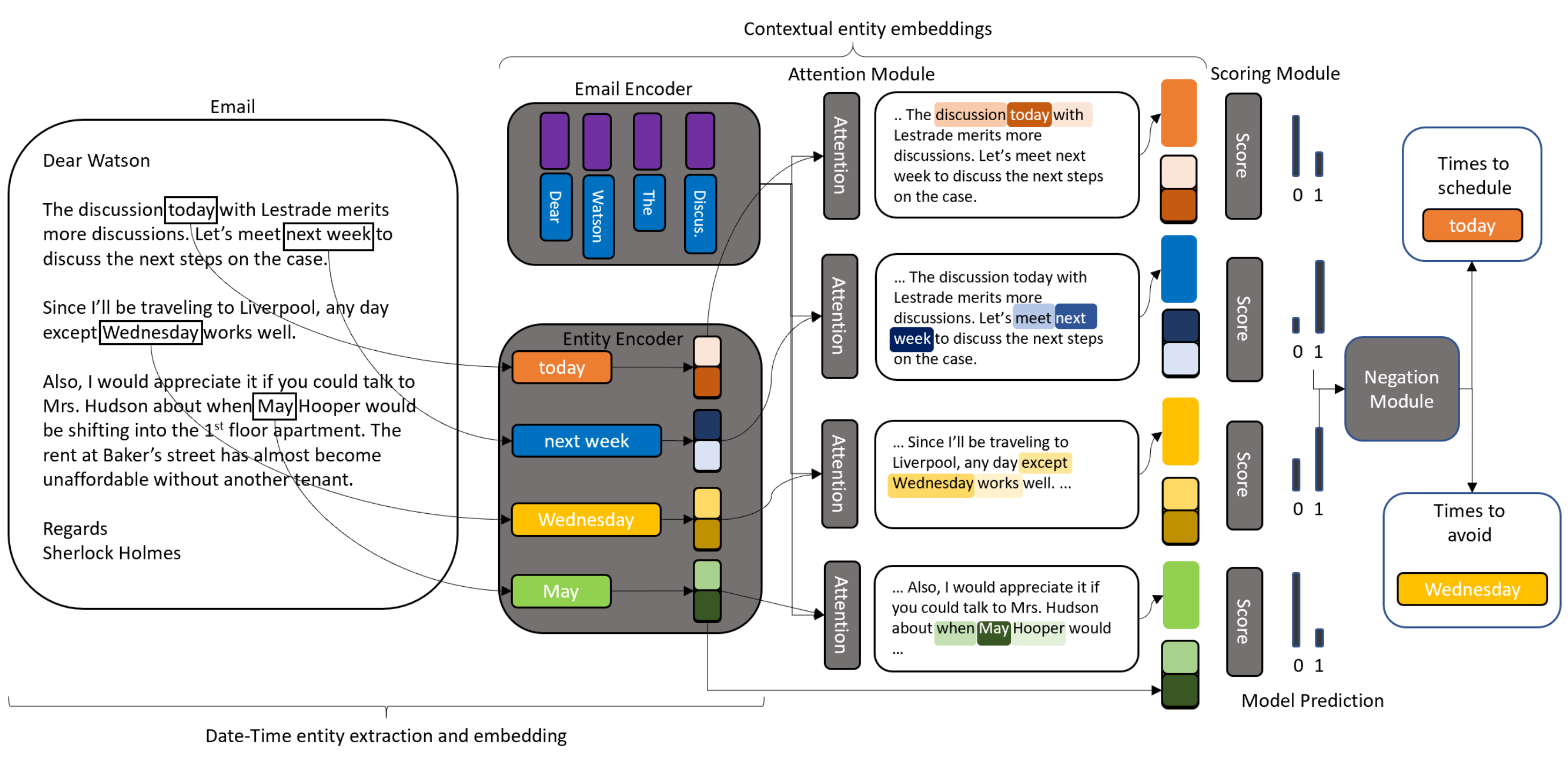}
    \caption{The 3 modules of SHERLOCK: First a high recall rule based extractor generates the potential entities. The neural module then takes the email and the entities and generates scores for each entity. Only the relevant entities are passed to the final negation module to detect times to schedule and times to avoid.}
    \label{fig:model}
\end{figure*}

In this paper, we address the task of recognizing date-time entities required by an AI scheduling assistant for correctly scheduling meetings. Cortana from Microsoft Scheduler, Clara from Clara Labs and Amy from X.ai are examples of such email based digital assistants for scheduling meetings. For such systems, a user organizing the meeting adds the digital assistant as a recipient in an email with other attendees and delegates the task of scheduling to the digital assistant in natural language. For the assistant to correctly schedule the meeting, it must correctly extract the date-time entities expressed by the user in the email to indicate the times they want the meeting scheduled, as well as the times that do not work for them. The verbose nature of emails often exacerbates the difficulty of identifying relevant date-time entities; since the number of distractor (i.e valid date-time entities not pertinent to the task) tend to increase (Eg: In Fig. \ref{fig:model} ``today'' serves as a distractor entity). 

To this end, we present SHERLOCK: {\bf S}c{\bf H}eduling {\bf E}ntity {\bf R}ecovery by {\bf LO}oking at {\bf C}ontextual {\bf K}nowledge, a novel model for detecting relevant date-time entities in the context of scheduling as well as identifying the entities associated with a negation constraint. SHERLOCK comprises of 3 modules for identifying the relevant entities as well as negation constraints associated with them: 

\paragraph{\bf Date-Time Extractor:}A high recall date-time entity extractor to identify all date-time entities in an email 

\paragraph{\bf Entity Relevance Scorer:}A neural model to classify each of the extracted entities as being relevant to scheduling or not by considering the context presented in the email. 

\paragraph{\bf Negation Detector:}A negation module to identify if there exists a negation constraint associated with each of the extracted relevant entities. \mbox{}\\

Fig. \ref{fig:model} illustrates each module: the entity extractor extracts ``today'', ``next week'', ``Wednesday'' and ``May''. Each of these entities is scored by the neural module, and only ``next week'' and ``Wednesday'' are identified as being relevant to scheduling. Finally, the negation module identifies that ``Wednesday'' has a negation constraint. While SHERLOCK focuses on the task of scheduling, we believe that a similar approach can be used to tackle the problem of extracting relevant date-time entities from documents for other tasks.

The contributions of this paper are as follows:

\paragraph{Task specific date-time extractor:} A novel method for combining conventional high recall rule-based model with a novel neural model for incorporating contextual information to identify relevant date-time entities for the task at hand.
\paragraph{Identifying negation constraints for temporal entities:} A heuristic negation module that helps identify negation constraints associated with time entities in the context of scheduling meetings. To the best of our knowledge, prior to this work, negation constraints associated with time-entity extraction have not been studied before.

We first present our proposed method for extracting time entities relevant to the task of scheduling a meeting in (\S \ref{section:approach}). Next, we describe our approach for identifying negation constraints associated with extracted entities in (\S \ref{section:negation}). In (\S \ref{section:experiments}), we describe our experimental setup and baselines. We discuss the results in (\S \ref{section:results}) and show that SHERLOCK helps improve performance both on the task of identifying relevant entities as well as identifying negation constraints. We then present the related work in (\S \ref{section:related}), and finally conclude in (\S \ref{section:conclusion}).

%% file: texfiles/approach.tex
\section{Contextual Date-Time Extraction}
\label{section:approach}

In order to correctly extract relevant temporal entities in the context of scheduling meetings from an email, we first extract potential entities using an off-the-shelf date-time entity extractor. Both the email and the extracted entities are then encoded using neural modules. For each extracted entity, we then generate a context embedding using the encoded email and the encoded entity. Both the contextual and encoded entity embedding are then used to predict if an entity is relevant or not. We describe each component in detail below:

\subsection{Entity Extraction and Encoding}
Given an email $\bm{X} = \{w_1 \cdots w_n\}$, we first use a rule-based tagger for extracting potential date-time entities from an email. Specifically, we use LUIS\footnote{\url{https://www.luis.ai/home}} \citep{williams2015fast} for extracting the entities. The model is recall heavy and identifies potential time utterances (Eg: in Figure \ref{fig:model}, LUIS detects ``today'', ``next week'', ``wednesday'', ``may''). We denote the extracted entities as $\mathcal{E} = \{e_1 \cdots e_m\}$, where $e_i = \{e_{i, 1} \cdots e_{i, l_i}\}$ represents the $i^{th}$ entity and $l_i$ denotes the length of $e_i$.

For each entity $e_i$, we generate an embedding $u_{e_i} \in \mathbb{R}^{d_e}$ (where $d_e$ denotes the entity embedding dimension) as follows:

\begin{equation}
    \label{eqn:word}
    \begin{aligned}
        t_{i, j} &= {\tt LookUp}(e_{i, j}) \\
        r_{i,j} &= {\tt CharEncoder}(e_{i, j}) \\
        h_{i, j} &= [r_{i,j} ; t_{i, j}] \\
        u_{e_i} &= {\tt Seq2SeqEncoder}(h_{i, 1} \cdots h_{i, l_i})
    \end{aligned}
\end{equation}
In Equation (\ref{eqn:word}), $t_{i, j}$ denotes the word level embedding of the $j^{th}$ word if the $i^{th}$ entity ($e_{i,j}$). As is standard practice, OOV words all share a common word embedding, while other entities encountered during training are represented by a learnt vector. We also augment this with an embedding from a character level encoder. $r_{i, j}$ denotes the word level embedding obtained by passing $e_{i,j}$\footnote{Technically the embeddings associated with characters of $e_{i,j}$ are passed to the character level encoder} through a character level encoder, which allows the model to represent OOV entities. The two embeddings are concatenated, and then passed through another \texttt{Seq2SeqEncoder} model (any Sequence-to-Sequence encoder \cite{seq2seq}) to get the final entity encoding $u_{e_i}$\footnote{For a unidirectional encoder, the final hidden state is used as the embedding. The concatenation of the forward and backward hidden states is used for a bidirectional encoder}.

\subsection{Contextual Entity Embeddings}
From Figure \ref{fig:model}, we can observe that it is clear from context that ``May'' is not a time entity, and ``today'' is not an entity relevant to scheduling. We want to capture this contextual information for each entity. To do so, we first encode the email as follows: 
\begin{equation}
    \label{eqn:email}
    \begin{aligned}
        (v_{w_1} \cdots v_{w_n}) = {\tt Seq2SeqEncoder}(w_1 \cdots w_n)
    \end{aligned}
\end{equation}

In Equation (\ref{eqn:email}), $v_{w_i}$ denotes the embedding for the $i^{th}$ word of the email $\bm{X}$. $d_w$ here denotes the embedding size for the email embedding.

Once we have the email embeddings, we then compute the contextual embedding for each entity using an attention mechanism \cite{bahdanau2014neural}. For entity $e_i$, given the entity embedding $u_{e_i}$, and the email embeddings $(v_{w_1} \cdots w_{w_n}), v_{w_i} \in \mathbb{R}^{d_w}$, the contextual embedding $c_{e_i} \in \mathbb{R}^{d_w}$ is obtained as follows:

\begin{equation}
    \label{eqn:attn}
    \begin{aligned}
        a_{w_j} &= A v_{w_j} + b \\
        b_{w_j} &= {\tt tanh}(a_{w_j} + u_{e_i}) \\
        logit_{w_j} &= B b_{w_j} + d\\
        \alpha{w_j} &= {\tt softmax}(logit_{w_j}) \\
        c_{e_i} &= \sum_{w^{\prime} \in \{w_1 \cdots w_n\}} \alpha_{w^{\prime}} v_{w^{\prime}}
    \end{aligned}
\end{equation}

Where $A \in \mathbb{R}^{d_e \times d_w}, b \in \mathbb{R}^{d_e}, B \in \mathbb{R}^{d_e \times 1}, d \in \mathbb{R}$ are learned parameters. The final entity embedding ($f_{e_i}$) is the concatenation of the entity embedding and the contextual embedding. Finally, for each entity, we generate a probability score to indicate if an entity is relevant or not.

\begin{equation}
    \label{eqn:score}
    \begin{aligned}
    f_{e_i} &= [u_{e_i}; c_{e_i}] \\
    s_{e_i} &= \sigma(Mf_{e_i} + g)
    \end{aligned}
\end{equation}

Where $M \in \mathbb{R}^{(d_e + d_w) \times 1}, g \in \mathbb{R}$ are learned parameters, and $\sigma$ indicates the sigmoid function.

\subsection{Learning}
Given the entities that are relevant to scheduling $\mathcal{Y} \subseteq \mathcal{E}$ (Eg: ``next week'' and ``Wednesday'' in Figure \ref{fig:model}), we train the model with a scoring loss as follows:
\begin{equation}
    \label{eqn:loss-score}
    \mathcal{L}_{s} = -(\sum_{e \in \mathcal{Y}} \log(s_e) + \sum_{e \in \mathcal{E} \setminus \mathcal{Y}} \log(1 - s_e))
\end{equation}

Similar to \cite{ruder2017overview, gehrmann-etal-2018-bottom, li-etal-2018-improving}, we find that augmenting the learning with a related auxiliary task helps improve performance. In this case, a simple related auxiliary function is the task of sequence tagging. Specifically, given the email $\bm{X}$, and the relevant entities $\mathcal{Y}$, we tag the location of each entity with I-Time tag, and every other token with an O tag. Let the generated tags be $\bm{z} = (z_1 \cdots z_n)$ and let $C$ denote the set of possible tagging labels (in our case 2: \{I-Time, O\}) We then train a standard CRF for tagging as follows:

\begin{equation}
    \label{eqn:loss-tagging}
    \begin{aligned}
        g_i &= P v_i + q \\
        {\tt score}(\bm{X}, \bm{z}) &= \sum_{i=2}^{n} T_{z_i, z_{i-1}} + \sum_{i=1}^{n} g_{i, z_i} \\
        p(\bm{z} | \bm{X}) &= \frac{e^{{\tt score}(\bm{X}, \bm{z})}}{\sum_{\bm{z}^{\prime} \in \mathcal{Z}}e^{{\tt score}(\bm{X}, \bm{z}^{\prime})}} \\
        \mathcal{L}_{t} &= -({\tt score}(\bm{X}, \bm{z}) - \\
            & \log(\sum_{\bm{z}^{\prime} \in \mathcal{Z}}e^{{\tt score}(\bm{X}, \bm{z}^{\prime})}))
    \end{aligned}
\end{equation}

Where $P \in \mathbb{R}^{d_w \times |C|}, q \in \mathbb{R}^{|C|}$ are trainable parameters, $T \in \mathbb{R}^{|C| \times |C|}$ is the transition matrix and $\mathcal{Z}$ is the set of all possible sequence labels.

The final loss that we optimize for is

\begin{equation}
    \label{eqn:loss-final}
    \mathcal{L}_{final} = \gamma \mathcal{L}_{s} + (1 - \gamma) \mathcal{L}_{t}
\end{equation}
Where $\gamma$ balances between the two loss functions.

\subsection{Choosing the Prediction Threshold}
In order to find the threshold for classifying the positive class (i.e $t$ such that $e_i = 1$ if $s_{e_i} > t$), we compute the F1 score on the validation set using a grid of thresholds\footnote{We use the \href{https://scikit-learn.org/stable/modules/generated/sklearn.metrics.precision_recall_curve.html}{precision\_recall\_curve} provided by sklearn \cite{sklearn_api}}, and choose the threshold maximizing the F1 score.

\section{Identifying Negation Constraints}
\label{section:negation}
\begin{figure*}[!thb]
    \centering
    \includegraphics[width=1.0\textwidth]{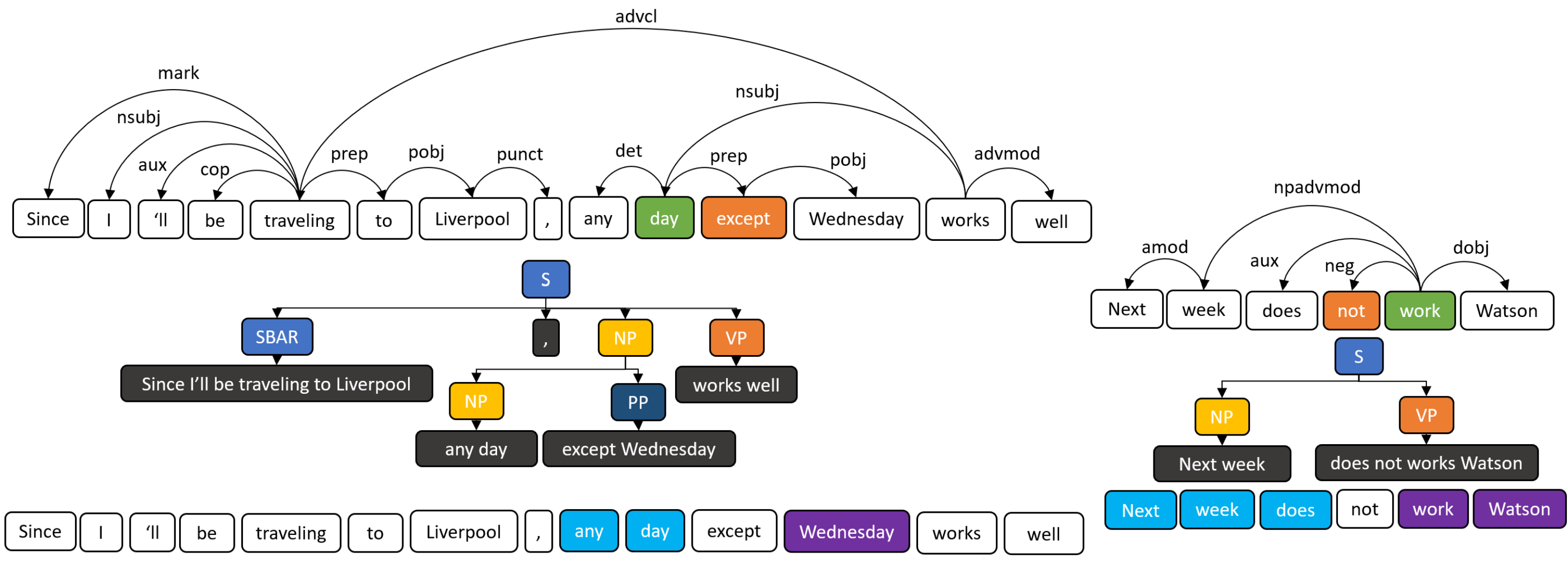}
    \caption{The negation extraction model. \textcolor{customOrange}{Orange} indicates the negation cue, \textcolor{customGreen}{Green} denotes the governing node. \textcolor{customPurple}{Purple} denotes the narrow scope, and \textcolor{customLightBlue}{Light Blue} denotes the wide scope.}
    \label{fig:negation}
\end{figure*}

For a Scheduling Assistant to be able to correctly schedule meetings, understanding negations is crucial; otherwise it can lead to an unsatisfactory user experience (E.g.: In Figure \ref{fig:model}, the meeting being scheduled on Wednesday would be a frustrating experience for the organizer Sherlock).
Only about 10\% of scheduling requests in our dataset have negation constraints. Building a model directly for the task did not show promising results from our preliminary experiments. We hypothesize this was due to the small volume of the data as well as the lack of good quality supervised data. Consequently, to find negated time-entities, we adopt the approach of first finding the negation scope. If an entity occurs inside the negation scope, we mark it to be negated.

In order to find the negation scope, we build on the approach proposed in \citet{rosenberg2013negation}. We first find the negation cue (``except'' in Figure \ref{fig:model}). To find the negation cue, we first tokenize the email into sentences. For each sentence, we try fo find if cue from a set of negating cues (Appendix \ref{app:negation_cues}) occurs in the sentence. 

After finding the negation cue, we identify the POS tag of the negating cue (Prep. for ``except''). Given the POS tag and the negation cue, we trigger a set of heuristics to identify the negation scope. Most heuristics work by identifying the negation cue from the dependency parse of the sentence as well as the governor of the negating word. Generating the narrow scope of negation (i.e. not containing the subject) then involves identifying the constituent from the constituency parse that contains both the negation cue and the governor word (``any day except Wednesday'', see Figure \ref{fig:negation}). This constituent is considered to be the candidate narrow scope, and usually, the part following the cue is considered to be the narrow scope. 

For some cases, the narrow negation scope is not enough to identify the time entity being negated. Consider the second example from Figure \ref{fig:negation}:

\indent {\bf Example:} {\it Next week does not work Watson.} \\
\indent {\bf Narrow:} {\it Next week does not [work Watson].} \\
\indent {\bf Wide:} {\it [Next week] [does] not [work Watson].}

For this case, the narrow scope is not enough to identify the entity being negated (``next week''). To find the wide scope, the heuristics leveraging the dependency path starting from the governor word are used. The main idea is to find the subject associated with the governor node, and extract that as the wide scope (``Next week''). Following the guidelines set by \citet{morante2012conandoyle}, we also include the aux dependency node in the wide scope (``does'').

We also expand the heuristic set presented in \citet{rosenberg2013negation}, adding the following rules:
\begin{itemize}
    \item If a Noun Phrase (NP) acting as an adverbial modifier acts as a subject to the governor, we include it in the wide scope (Figure \ref{fig:negation})
    \item If a NP exists as a subject of a passive clause, we include it in the wide scope, as well as the passive auxiliary associated with it.
    \item A Prepositional Phrase (PP) acting as a subject to the governor is included in the wide scope.
    \item For the narrow scope, we prune out the subtree that exists as an object an adverbial clause relation (advcl) headed by the governor node.
\end{itemize}
Due to space constraints, we include examples for the above in Appendix \ref{app:negation_heuristics}.

After obtaining the narrow and wide scopes, we check if any entities are found in the narrow scope. If found, those entities are scored negated. If no entities are found in the narrow scope, we then check the wide scope to find negated entities. 

Finally, for some cases, we also use domain specific cues that imply a non-availability (For example, in ``Dr. John out of office on Monday.'', ``out of office'' implies an unavailability to meet.) When such implied negation cues are encountered, we default to a custom heuristic which marks any entity occurring within the sentence containing the cue word as a negation.

%% file: texfiles/experiments.tex
\section{Experimental Setup}
\label{section:experiments}
We first show the effectiveness of our proposed entity scoring method of incorporating context for improving temporal entity extraction on the TempEval-3 dataset \cite{uzzaman2013semeval} (\S\ref{subsec:public}). We then show the efficacy of SHERLOCK for the task of extracting the correct temporal entities relevant for the context of scheduling, a task for which context becomes substantially more important (\S\ref{subsec:datetime}). Finally, we show that SHERLOCK's negation module outperforms baseline methods on the task of identifying the entities with negation constraints (\S\ref{subsec:negation}). All our models have been implemented using the AllenNLP framework \cite{Gardner2017AllenNLP}. The hyperparameters for all the experiments can be found in Appendix \ref{app:hyperparams} and \ref{app:hyperparams-negation}.

\subsection{TempEval-2013}
\label{subsec:public}
\subsubsection{Dataset}
We use the TimeBank dataset \cite{pustejovsky2003timebank} which serves as the benchmark dataset for the TempEval series. The dataset consists of 256 documents, comprising of 95,391 tokens and 1,822 TimeEx entities for training and validation purposes, and 20 documents (6,375 tokens, 138 TimeEx) for serving as the test set.

\subsubsection{Baseline Models}
We show the performance of augmenting 3 rule-based models with our proposed model. Specifically, we consider SUTime \cite{chang2012sutime}, HeidelTime \cite{strotgen2010heideltime} and Syntime \cite{zhong2017time} as the rule-based extractors. We also compare against UWTime \cite{lee2014context}, a learning based model.

\subsubsection{Evaluation}
We use the official TempEval-3 scoring script and report the standard metrics. Specifically, we report the detection precision, recall and F1 with the relaxed and strict metrics. A gold mention is considered for the relaxed metric if any of the output candidates overlap with it and for the strict case, an exact string match is considered.

\subsection{Date-Time extraction for Scheduling}
\label{subsec:datetime}
This task aims at extracting the date-time entities necessary for the Scheduling Agent to correctly schedule the meeting. The task necessarily needs the model to incorporate context for making the correct prediction (E.g.: In Figure \ref{fig:model}, ``today'' is a valid date-time entity, but not relevant for scheduling, while ``May'' refers to a person.)
\subsubsection{Dataset}
We use an internal scheduling dataset for training and evaluating the models. The dataset consists of emails and annotated times to schedule. The training and validation set consists of 44,214 emails (4,589,631 tokens, and 48083 entities), while the test set consists of 4914 emails (519,021 tokens, 5233 entities).

\subsubsection{Baseline Models}
We compare the performance of our model against SUTime, HidelTime and LUIS. We use LUIS as our base date-time extractor since it provides a much larger coverage for date-time entities \footnote{For example, LUIS recognizes military time (``1530''), and has a much larger coverage for holidays}. 

\subsubsection{Evaluation}
We use the Strict F1 measure to compare the performance of the different models proposed.

\subsection{Negation Detection}
\label{subsec:negation}
Finally, we compare the performance of our proposed model on the task of negation extraction.

\subsubsection{Dataset}
We use an internal dataset for comparing different models on the task of negation extraction. The dataset consists of 1253 emails for which time-entities that are relevant to scheduling are selected, and those that are a part of a negation constraint are marked as negated entities. There exist 3231 time-entities, of which 1589 are negated entities.

\subsubsection{Baselines}
We compare our proposed method against a naive heuristic method as well as a neural model trained on a publicly available negation scope detection dataset.

{\bf Heuristic:} A naive heuristic model. If a negation cue is identified in a sentence, the model predicts that all entities in that sentence are negated.

{\bf NegNN:} We use a NegNN model \cite{fancellu2016neural}, modified to use BERT contextual embeddings and trained on the *SEM2012 Shared task \cite{morante2012sem}. The training, development and test sets are a collection of stories from Conan Doyle's {\it Sherlock Holmes}, with the cue and scope annotated. An entity is considered negated if it is a part of a negated scope, as predicted by the model. The performance of the modified NegNN model on the *SEM2012 Task can be found in Appendix \ref{app:negation}.

\subsubsection{Evaluation}
We measure the performance of different models by comparing the predicted set of negated entities and the gold labels for the entities. If the model makes a mistake (i.e. it predicts an entity to be negated, when it's not), that's considered a false positive. Likewise, any negated entities missed by the model contribute to the false negatives. We thus report the precision, recall and F1 score.

%% file: texfiles/results.tex
\section{Results and Analysis}
\label{section:results}
\subsection{TempEval-2013}
\input{tables/public_datasets.tex}

Table \ref{tab:public} shows the performance of SHERLOCK's entity scoring module on the TempEval-2013 dataset. Note that SHERLOCK is limited by the recall of the base rule-based extractor\footnote{The model only scores the predictions of the base extractor}. We observe that augmenting the rule-based model with SHERLOCK improves the precision for all three cases without a substantial drop in recall. Furthermore, the precision obtained for all the augmented models compares favorably with UWTime.

\subsection{Date-Time extraction for Scheduling}
\input{tables/internal.tex}
Table \ref{tab:internal} shows the performance of SHERLOCK for the Scheduling related date-time extraction task. As can be seen, being able to incorporate context yields a substantial improvement over the baseline methods.

We also observed that incorporating the tagging loss $\mathcal{L}_t$ helped improve performance (SHERLOCK vs SHERLOCK - $\mathcal{L}_t$). On investigating further, we observed that the attention weights associated with an entity for a model trained with $\mathcal{L}_t$ concentrated much better around the position of the entity in the email than for the model without it. To see why that is advantageous, consider the following example: 

\textit{``Let's schedule for tomorrow. Next month, I plan on taking up Mr Baskerville's case''}

Here, the model without $\mathcal{L}_t$ generates high attention weights for embeddings associated with ``tomorrow'', since the localization of the attention weights is much more spread out. Consequently, it also uses the embeddings associated with ``tomorrow'' for predicting the label of ``next month'', and hence, predicts it to be relevant to scheduling when it is not. Due to space constraints, we include our localization experiments in Appendix \ref{app:localization}.

\subsection{Negation Detection}
\input{tables/negation}
\input{tables/neg_ablation}

Table \ref{tab:negation} shows the performance of SHERLOCK compared to the baseline methods. 
We hypothesize the reason why SHERLOCK and the simple heuristic model outperform the neural baseline is two-fold: the neural negation model was trained on a dataset of Sherlock Holmes stories and consequently does not adapt well when used for negation extraction for emails; and that the neural model has no notion of implied negations.

To test this hypothesis, we split the negations into two categories: explicit negations (defined as a negation where the cue is one of the explicit negation cues), and the case wherein the negation is implied (any case that was not explicit was deemed implied). 50\% of emails in the negation dataset contained explicit negations only, 48\% contained implied negations only and 2\% contained both.

Table \ref{tab:neg_ablation} shows the performance of SHERLOCK and the baselines for both the explicit negation and the implied negation cases. Unsurprisingly, we see that both the baselines as well as SHERLOCK perform better on explicit negations than they do on implied negations. However, the gains observed by both the heuristic model and SHERLOCK substantially outperform NegNN, with SHERLOCK substantially outperforming the heuristic. Examples 1 and 2 in Table \ref{tab:negation-errors} give qualitative examples of where SHERLOCK outperforms the heuristic. 

\input{tables/negation_error_analysis.tex}

The primary source of errors for detecting implied negations is from failing to identify the correct cue. Since heuristics for implied negations are more heavily focused on precision, the absence of negation cues results in the model not detecting the implied negation, which in turn negatively impacts the recall. Examples 3, 4 and 5 in Table \ref{tab:negation-errors} show cases where the cue is not present in the heuristic set of implied cues.

For explicit negations, one source of errors is due to entity co-referencing. Consider Example 6: the negated time instance Tuesday is referenced as ``then'' and hence the negation scope ``then'' is insufficient to identify the correct negated entity. A few errors also stem from inherent ambiguity: in Example 7, the request can either be interpreted as being for anytime next week except Thursday 10am, or for 10 am on all days except Thursday. Finally, we also observe errors due to double negations (Example 8) and due to incorrect constituency and dependency parses.

%% file: tables/public_datasets.tex
\begin{table}[!thb]
\small
\centering
\begin{tabular}{@{}l@{}cccccc@{}}
\toprule
\multirow{2}{*}{\bf Model} & \multicolumn{3}{c}{\bf Strict} & \multicolumn{3}{c}{\bf Relaxed}  \\
\cmidrule{2-7}
 & {\bf Pre.} & {\bf Rec.} & {\bf F1} & {\bf Pre.} & {\bf Rec.} & {\bf F1} \\
 \midrule
SUTime & 80.0 & 81.2 & 80.6 & 90.0 & 91.3 & 90.7 \\
SUTime(+) & 85.9 & 79.7 & 82.7 & 93.6 & 87.0 & 90.2 \\
\midrule
HeidelTime   & 83.9 & 79.7 & 81.7 & 93.1 & 88.4 & 90.7 \\
{HeidelTime(+) }  & 84.6 & 79.7 & 82.1 & 93.1 & 87.7 & 90.3 \\
\midrule
Syntime   & 91.4 & 92.7 & 92.1 & 94.3 & 95.7 & 95.0 \\
{Syntime(+) }  & 92.7 & 92.0 & 92.4 & 94.9 & 94.2 & 94.6 \\
\midrule
UWTime & 84.6 & 83.4 & 84.0 & 92.8 & 91.5 & 92.1 \\
\bottomrule
\end{tabular}
\caption{Performance on TempEval Dataset. Models with (+) indicate that the base extractor is augmented with the entity scoring module (Scale: 0-100)}
\label{tab:public}
\vspace{-2mm}
\end{table}

%% file: tables/internal.tex
\begin{table}[!thb]
\small
\centering
\begin{tabular}{@{}lccr@{}}
\toprule
{\bf Model} & {\bf Precision} & {\bf Recall} & {\bf F1}  \\
\midrule
LUIS   & 0.38 & 0.98 & 0.54 \\
SUTime   & 0.59 & 0.79 & 0.68 \\
HeidelTime & 0.66 & 0.86 & 0.75 \\
SHERLOCK - $\mathcal{L}_t$ & 0.91 & 0.96 & 0.93 \\
SHERLOCK & 0.91 & 0.98 & 0.94 \\
\bottomrule
\end{tabular}
\caption{Performance on Date-Time Extraction for Scheduling. SHERLOCK - $\mathcal{L}_t$ denotes the SHERLOCK model without the tagging loss (Scale: 0 - 1)}
\label{tab:internal}
\vspace{-2mm}
\end{table}

%% file: tables/negation.tex
\begin{table}[!thb]
\small
\centering
\begin{tabular}{@{}lccr@{}}
\toprule
{\bf Model} & {\bf Precision} & {\bf Recall} & {\bf F1} \\
\midrule
NegNN & 0.73 & 0.13 & 0.22 \\
Heuristic & 0.78 & 0.63 & 0.70 \\
SHERLOCK & 0.91  & 0.62 & 0.74 \\
\bottomrule
\end{tabular}
\caption{Negation Performance (Scale: 0 - 1)}
\label{tab:negation}
\vspace{-2mm}
\end{table}

%% file: tables/neg_ablation.tex
\begin{table}[!thb]
    \small
    \centering
    \begin{tabular}{@{}lcccr@{}}
    \toprule
    {\bf Category} & {\bf Model} & {\bf Precision} & {\bf Recall} & {\bf F1} \\
    \midrule
    \multirow{3}{*}{Explicit} & NegNN & 0.83 & 0.25 & 0.39 \\
    & Heuristic & 0.76 & 0.86 & 0.81 \\
    & SHERLOCK & 0.94  & 0.87 & 0.90 \\
    \midrule
    \midrule
    \multirow{3}{*}{Implied} & NegNN & 0.23 & 0.01 & 0.03 \\
    & Heuristic & 0.83 & 0.42 & 0.56 \\
    & SHERLOCK & 0.87  & 0.40 & 0.54 \\
    \bottomrule
    \end{tabular}
    \caption{Explicit vs implied negations (Scale: 0 - 1)}
    \label{tab:neg_ablation}
    \vspace{-2mm}
\end{table}

%% file: tables/negation_error_analysis.tex
\begin{table*}[!htb]
    \small
    \centering
    \begin{tabular}{@{}l@{}lccr@{}}
    \toprule
    {\bf Idx} & \multicolumn{1}{c}{{\bf Example}} & {\bf Heuristic} & {\bf SHERLOCK} & {\bf Correct} \\
    \midrule
    1 & \makecell[l]{Mycroft cannot do Monday but Tuesday \\should work fine.} & [Monday, Tuesday] & [Monday] & [Monday] \\
    \midrule
    2 & If Watson is not busy, Wednesday also works. & [Wednesday] & [] & []\\
    \midrule
    3 & I'm slammed on Thursday. - Lestrade & [] & [] & [Thursday] \\
    \midrule
    4 & \makecell[l]{I am out of town on Wednesday Irene but \\Thursday might work.} & [] & [] & [Wednesday] \\
    \midrule
    5 & \makecell[l]{I am completely booked with \\appointments on Thursday Sherlock.\\-- Watson} & [] & [] & [Thursday] \\
    \midrule
    6 & \makecell[l]{Mr. Holmes, my trip's on Tuesday.\\I really can't meet then.} & [] & [] & [Tuesday] \\
    \midrule
    7 & \makecell[l]{Let's just meet next week any day except\\Thursday at 10:00 am. - Holmes} & \makecell[c]{[next week,\\Thursday at 10:00 am]} & [Thursday] & \makecell[r]{[Thursday at 10:00 am]} \\
    \midrule
    8 & \makecell[l]{Next week would not be possible, except\\on Friday.} & [Next week, Friday] & [Next week, Friday] & [next week] \\
    \bottomrule
    \end{tabular}
    \caption{Examples of SHERLOCK's Negation Model's predictions and errors}
    \label{tab:negation-errors}
    \vspace{-2mm}
\end{table*}

%% file: texfiles/relatedWork.tex
\section{Related Work}
\label{section:related}
Existing approaches for time expression extraction can be categorized into rule-based methods and learning-based methods.

\textbf{Rule-based Methods} Rule-based methods like HeidelTime, and SUTime mainly handcraft deterministic rules to identify time expressions. TempEx and GUTime use both hand-crafted rules and machine-learnt rules to resolve time expressions \cite{mani2000robust, verhagen2005automating, blamey2013the}. HeidelTime manually designs rules with time resources to recognize time expressions \cite{strotgen2010heideltime}. SUTime designs deterministic rules at three levels (i.e., individual word level, chunk level, and time expression level) for time expression recognition \cite{chang2012sutime}. A recent type-based time tagger, SynTime, designs general heuristic rules with a token type system to recognize time expressions \cite{zhong2017time}. TOMN \cite{zhong2018time} uses the token regular expressions, similar to SUTime \cite{chang2012sutime} and SynTime \cite{zhong2017time}, and further groups them into three token types, similar to SynTime. TOMN also leverages statistical information from entire corpus to improve the precisions and alleviate the deterministic role of deterministic and heuristic rules.

\textbf{Learning-based Method} Learning-based methods in TempEval series mainly extract features from text (e.g., character features, word features, syntactic features, and semantic features), and on the features apply statistical models (e.g., CRFs) to model time expressions \cite{Bethard_thecleartk-timeml, filannino2013mantime, llorens2010tipsem, uzzaman2010trips}. Besides the standard methods, \cite{Angeli12parsingtime, Angeli_language-independentdiscriminative} exploit an EM-style approach with compositional grammar to learn latent time parsers. \cite{lee2014context} leverage a learnt CCG \cite{steedman1996surface} parser and define a lexicon with linguistic context to model time expressions, using the loose structure information by grouping the constituent words of time expression under three token types.

\textbf{Negation Scope Detection:} Most negation detection research has focused in the Bio-Medical domain \cite{mehrabi2015deepen, agarwal2010biomedical}. Non Bio-Medical text related negation detection tasks usually involve learning supervised classifiers over hand-crafted features leveraging syntactic structure (constituency and dependency parses) \cite{velldal2012speculation, lapponi2012uio, chowdhury2012fbk, white2012uwashington, abu2012umichigan}. The current state of the art learned method uses a Neural BiLSTM-CRF model \cite{fancellu2016neural}. However, the corpus available for negation detection is on Sherlock Holmes stories (*SEM2012 Shared task \cite{morante2012conandoyle}), and consequently, as shown in this work, do not adapt well on language used in other document styles (like emails). In this work, we built over the work of \cite{rosenberg2013negation}, who develop linguistic rules over constituency and dependency parses to identify negation scopes. The primary advantage of leveraging their work is that it is not strongly tied to the *SEM 2012 dataset, and we found this to generalize better. 

Finally, there has been some work on directly training a model to extract entities and associated negation constraints \citep{bhatia-etal-2019-joint}. However, these works usually assume the availability of good quality annotated negated entities. Given enough annotated data, exploring this direction would be an interesting line of future work.

%% file: texfiles/conclusion.tex
\section{Conclusion}
\label{section:conclusion}

In this paper, we presented a novel model that leverages conventional high recall rule-based models and neural models for utilizing contextual information for identifying task relevant temporal entities. Our proposed model, when used in conjunction with 3 different rule-based models, achieves substantial precision gains for all of them without suffering from a huge recall drop. Further, the model substantially outperforms baseline methods for the task of identifying relevant date-time entities for the task of scheduling a meeting. 

We also presented a novel approach for identifying the negation constraints of date-time entities. Identifying the negation constraints associated with date-time entities correctly is necessary for the task of scheduling. We showed that the existing neural approaches for detecting negation scopes do not transfer well, and that our proposed model based on heuristics defined over constituency and dependency parses achieves strong performance gains, especially for the case of explicit negations.

%% file: texfiles/acknowledgements.tex
\section*{Acknowledgements}
We would like to thank Vishwas Suryanarayanan for his valuable comments and discussions. We would also like to thank the anonymous reviewers, whose valuable comments and suggestions helped shape the paper into its current form.

%% file: texfiles/appendix.tex
\section{Negation Cues}
\label{app:negation_cues}
We use the following cues for detecting explicit negations scopes:
\begin{itemize}
    \item[] n't (This includes all words that get tokenized with this suffix like won't, can't etc.)
    \item[] not
    \item[] never
    \item[] neither
    \item[] nor
    \item[] no
    \item[] nothing
    \item[] nobody
    \item[] instead of
    \item[] without
    \item[] rather than
    \item[] failed to
    \item[] avoid
    \item[] other than
    \item[] unable
    \item[] negative
    \item[] except
    \item[] none
\end{itemize}
There also exist implied negations that indirectly imply the unavailability of the user. We list them below:
\begin{itemize}
    \item[] out of office
    \item[] ooo (the commonly used abbreviation for out of office)
    \item[] out of facility
    \item[] oof (the commonly used abbreviation for out of facility)
    \item[] vacation
    \item[] personal time off
    \item[] pto (the commonly used abbreviation for personal time off)
    \item[] busy
\end{itemize}

\section{Negation Heuristics}
\label{app:negation_heuristics}
We describe the modifications made to the model presented in \cite{rosenberg2013negation}.

\subsection{Additional Rules: Wide scope}
Instead of considering just the words connected by a \textit{nsubj} relation (both in the general case as well as when the governor is linked by a \textit{conj} link, in which case the subject is identified bt the term in the first coordinate clause), we include words connected by the \textit{nsubjpass} as well as the \textit{npadvmod} link. Further, we include a word connected by an auxpass link if it is directly attached with the governor word.

\indent {\bf Example:} {\it Next week does not work Mycroft.} \\
\indent {\bf Relevant Dependency Relations: } neg(work, not), npadvmod(work, week) \\
\indent {\bf Expected Scope:} {\it [[Next week], [does], [work Mycroft]]}

If we fail to account for the npadvmod relation, the wide scope does not include [Next week].

\indent {\bf Example:} Watson was not amused by Sherlock's antics.\\
\indent {\bf Relevant Dependency Relations:} nsubjpass(amused, Watson), neg(amused, not) \\
\indent {\bf Expected Scope:} [[Watson], [was], [amused]]

Likewise, in this case, if we fail to account for the nsubjpass node, then we don't account for the wide scope [Watson]. Similarly, if we fail to account for the auxpass node, we miss out on the wide scope [was].

Finally, another case where we need to expand the wide scope is if a Prepositional Phrase acts as a subject for the governor. Consider the following:

\indent {\bf Example:} Before Wednesday does not work.\\
\indent {\bf Relevant Dependency Relations:} prep(work, Before), pobj(Before, Wednesday), neg(work, neg) \\
\indent {\bf Expected Scope:} [[Before Wednesday], [does], [work]]

Here, the prepositional phrase (Before Wednesday) acts as a subject to the governor node, and consequently needs to be included in the wide scope.

\subsection{Additional Rules: Narrow Scope Pruning}
While computing the narrow scope, if the scope contains a word that forms an adverbial clause relation with the governor, we remove the dependency subtree associated with the advcl word. We give an example of such a case below:

\indent {\bf Example:} {\it This won't work as Mary is free next week.} \\
\indent {\bf Original Narrow Scope:} {\it [work, as, Mary, is, free, next, week]} \\
\indent {\bf advcl subtree:} {\it [as, Mary, is, free, next, week]} \\
\indent {\bf Proposed Narrow scope:} {\it [work]}

\section{Hyperparameters: Contextual Date-Time Extraction}
\label{app:hyperparams}
For both TempEval-2013 task and the Date-Time extraction for Scheduling task, we use the same model. The hyperparameters used are as follows:

\subsection{Entity Extraction and Encoding} 
\begin{itemize}
    \item[] \texttt{CharEncoder:} CNN Model similar to \cite{ma2016end}
    \begin{itemize} 
        \item[] Activation: ReLU
        \item[] Num Filters: 128
        \item[] Input Embedding Size: 16  
    \end{itemize}
    \item[] \texttt{LookUp:} Embedding Size: 50
    \item[] \texttt{Seq2SeqEncoder:} 
    \begin{itemize} 
        \item[] Type: Bidirectional GRU \cite{gru}
        \item[] Input Size: 128 + 50 = 178
        \item[] Hidden Size: 64
        \item[] Number of Layers: 1   
    \end{itemize}
\end{itemize}
\subsection{Contextual Entity Embedding}
\begin{itemize}
    \item[] \texttt{Seq2SeqEncoder:}
    \begin{itemize}
        \item[] Embeddings: ELMo embeddings \cite{elmo}
        \item[] Encoder:
        \begin{itemize}
            \item[] Type: Bidirectional GRU
            \item[] Input Size: 1024 \textit{// ELMo Size}
            \item[] Hidden Size: 64
            \item[] Dropout: 0.5
            \item[] Number of Layers: 2  
        \end{itemize}
    \end{itemize} 
\end{itemize}

Total Number of parameters: 96962040

\subsection{Loss Weighing Factor}
We use $\gamma=0.99$
\subsection{Training Hyperparameters}
Each model was trained for 75 epochs with a patience of 20, using Adam \cite{kingma2014adam} as the optimizer.

\input{tables/semeval2012_negation}

\subsection{Hyperparameter Search}
All models were trained on a single K80 instance. Contextual Date-Time experiments take about 2 days to run, while TempEval experiments take 6 hours to run. Hyperparameter search was carried out on the learning rates using a grid search in the range of [1e-3, 1e-2].
Tuning other parameters did not yield substantial gains.

\section{Hyperparameters: Negation Model}
\label{app:hyperparams-negation}
We reimplement the heuristics outlined in \cite{rosenberg2013negation} as well as the additional heuristics mentioned in the paper. We use the following components for our reimplementation:
\begin{itemize}
    \item[] {\bf POS Tagger:} Spacy POS Tagger\footnote{\href{https://spacy.io/usage/linguistic-features/}{Spacy POS Tagger}} using the "en\_core\_web\_sm" model.
    \item[] {\bf Constituency Parser:} The Berkeley Neural Parser Model \cite{Kitaev-2018-SelfAttentive}.\footnote{\href{https://github.com/nikitakit/self-attentive-parser}{Berkeley Neural Parser}}
    \item[] {\bf Dependency Parsers:} We use an ensemble of two parsers
    \begin{itemize}
        \item[] The AllenNLP implementation of the model presented in \cite{dozat2016deep}\footnote{\href{https://storage.googleapis.com/allennlp-public-models/biaffine-dependency-parser-ptb-2020.04.06.tar.gz}{AllenNLP Biaffine Dependency Parser}}
        \item[] The Spacy "en\_core\_web\_sm" model\footnote{\href{https://explosion.ai/demos/displacy}{Spacy Dependency Parser}}
    \end{itemize}
\end{itemize}

\section{Neural Model Training on Sherlock Dataset}
\label{app:negation}
We experiment with two models on the *SEMEval 2012 dataset, which yield comparable performance to the model presented by \cite{fancellu2016neural}. 

We experiment with tuning a BERT model \cite{devlin2018bert}. Specifically, we compare the performance of the "bert\_base\_cased\_en" and "bert\_base\_multilingual\_cased" model\footnote{\href{https://github.com/google-research/bert}{BERT Models}}. We use the HuggingFace Transformers implementation of the models \cite{Wolf2019HuggingFacesTS}. Specifically, we try fine-tuning BERT as well as using BERT as a feature extractor, and using an LSTM over the BERT model. We observed that using an LSTM performed a little better (Table \ref{tab:negation-semeval}), and consequently use that model for the experiments presented in the paper.

\section{Better Localization for models trained with $\mathcal{L}_t$}
\label{app:localization}

\begin{figure}[!thb]
    \centering
    \includegraphics[width=0.48\textwidth]{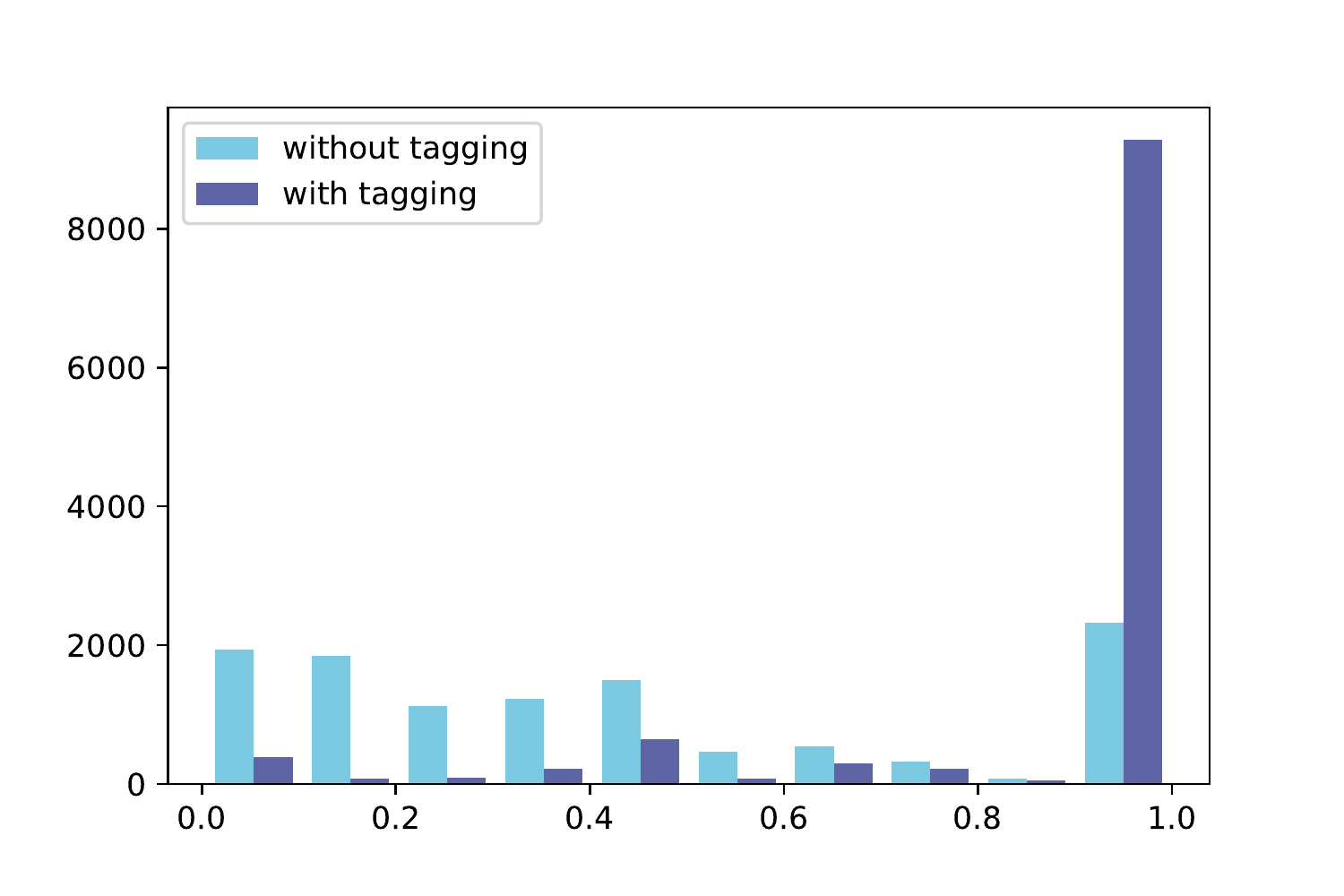}
    \caption{Overlap Score: With and without $L_{tagging}$}
    \label{fig:model-ablation}
\end{figure}

As mentioned in the paper, we also observed that the model augmented with $\mathcal{L}_{tagging}$ has an improved localization of entities by the attention module. Specifically, given an entity and the context, the attention weights induced by the entity (Equation \ref{eqn:attn}) are localized more around the entity text for the case of the augmented model \footnote{For example: ``Let us meet next week.'', a good localization of the attention weights for entity ``next week'' would be if the attention weights were high around ``next week'' in text.} In order to validate this hypothesis, we cluster the attention weights generated by an entity into two clusters using KMeans++ \cite{kmeans++} with $k=2$. We then extract the tokens associated with the smaller cluster, and consider that to be the localized context (since these weights are the highest, the embeddings associated with these tokens have the maximum impact in terms of predicting if the entity is relevant or not). For the localized context, we measure the degree of overlap with the original entity\footnote{Given a localized context $\{c_1 \cdots c_k\} \subset \mathbb{X}$ for the entity $e_i$, we compute $\frac{|\{c_1 \cdots c_k\} \cap e_i|}{k}$ } as a measure of quality of localization.

Figure \ref{fig:model-ablation} shows the histogram of the coverage of a model trained with and without $\mathcal{L}_{tagging}$. As can be seen, the model with the tagging loss is much better at concentrating the attention weights around the entity in consideration. To see why this is advantageous, consider the following example: 

\textit{``Let's schedule for tomorrow. Next month, I plan on taking up Mr Baskerville's case.''} \\
{\bf Entity in consideration:} Next month \\
{\bf With tagging: } Next month \\
{\bf Without tagging:} tomorrow . next month

Here, the model without tagging also uses the embeddings associated with tomorrow for predicting the label of next month, and consequently, predicts it to be relevant to scheduling when it is not.

%% file: tables/semeval2012_negation.tex
\begin{table}[!htb]
    \small
    \centering
    \begin{tabular}{@{}l@{}|@{}c@{}c@{}c@{}|@{}c@{}c@{}r@{}}
    \toprule
    \multirow{2}{*}{{\bf Model}} & \multicolumn{3}{c}{{\bf Tag Level}} & \multicolumn{3}{c}{{\bf Exact Match}} \\
    \cmidrule{2-7} 
    & { \bf Pre. } & { \bf Rec. } & { \bf F1 } & { \bf Pre. } & { \bf Rec. } & { \bf F1 } \\
    \midrule
    {Fancellu et al } & 0.92 & 0.85 & { 0.88 } & 0.99 & 0.64 & 0.77 \\
    {BERT Ft } & 0.83 & 0.89 & 0.86 & 0.73 & 0.82 & 0.73 \\
    {BERT + LSTM } & 0.92 & 0.87 & 0.89 & 0.73 & 0.78 & 0.76 \\
    \bottomrule
    \end{tabular}
    \caption{*SEMEval 2012 Model performance. Ft indicates fine-tuned}
    \label{tab:negation-semeval}
    \vspace{-2mm}
\end{table}